\UseRawInputEncoding
\documentclass[conference]{IEEEtran}
\IEEEoverridecommandlockouts
\usepackage{cite}
\usepackage{amsmath,amssymb,amsfonts}
\usepackage{algorithmic}
\usepackage{graphicx}
\usepackage{textcomp}
\usepackage{xcolor}
\usepackage{multirow}
\usepackage{tikz}
\usepackage{fancyhdr}

\fancypagestyle{}

\fancypagestyle{}
{
    
    \lhead{\footnotesize THIS PAPER HAS BEEN ACCEPTED AT \underline{\textit{ICMLA 2022}} AND THE FINAL VERSION HAS BEEN SUBMITTED}
    \rhead{\footnotesize \thepage}
    \cfoot{}
}
\def\BibTeX{{\rm B\kern-.05em{\sc i\kern-.025em b}\kern-.08em
    T\kern-.1667em\lower.7ex\hbox{E}\kern-.125emX}}

\usepackage{comment}

\usepackage{amsmath}
\usepackage{graphicx}
\usepackage[colorlinks=true, allcolors=blue]{hyperref}

\usepackage{color}
\definecolor{Orange}{rgb}{1,0.5,0}
\newcommand{\todo}[1]{\textsf{\textbf{\textcolor{Orange}{[#1]}}}}

\title{Using Natural Language Processing to Predict Costume Core Vocabulary of Historical Artifacts
\thanks{\textsuperscript{1} Amr Hilal is also affiliated with the Department of Computer and Systems Engineering, Alexandria University, Egypt.}
}

\author{\IEEEauthorblockN{Madhuvanti Muralikrishnan}
\IEEEauthorblockA{\textit{Computer Science Dept.} \\
\textit{Virginia Tech}\\
Blacksburg, VA \\
madhuvantim@vt.edu}
\and
\IEEEauthorblockN{Amr Hilal \textsuperscript{1}}
\IEEEauthorblockA{\textit{Informatics Lab} \\
\textit{Virginia Tech}\\
Blacksburg, VA \\
ahilal@vt.edu}
\and
\IEEEauthorblockN{Chreston Miller}
\IEEEauthorblockA{\textit{Informatics Lab} \\
\textit{Virginia Tech}\\
Blacksburg, VA \\
chmille3@vt.edu}
\and
\IEEEauthorblockN{Dina Smith-Glaviana}
\IEEEauthorblockA{\textit{Apparel, Housing, 
and Resource Management Dept.} \\
\textit{Virginia Tech}\\
Blacksburg, VA \\
dinacs@vt.edu}
}

\begin{document}
\maketitle
\thispagestyle{fancy}
\begin{abstract}

Historic dress artifacts are a valuable source for human studies. In particular, they can provide important insights into the social aspects of their corresponding era. These insights are commonly drawn from garment pictures as well as the accompanying descriptions and are usually stored in a standardized and controlled vocabulary that accurately describes garments and costume items, called the Costume Core Vocabulary. Building an accurate Costume Core from garment descriptions can be challenging because the historic garment items are often donated, and the accompanying descriptions can be based on untrained individuals and use a language common to the period of the items. In this paper, we present an approach to use Natural Language Processing (NLP) to map the free-form text descriptions of the historic items to that of the controlled vocabulary provided by the Costume Core. Despite the limited dataset, we were able to train an NLP model based on the Universal Sentence Encoder to perform this mapping with more than 90\% test accuracy for a subset of the Costume Core vocabulary. We describe our methodology, design choices, and development of our approach, and show the feasibility of predicting the Costume Core for unseen descriptions. With more garment descriptions still being curated to be used for training, we expect to have higher accuracy for better generalizability. 
\end{abstract}

\section{Introduction}

Throughout history, various terms have been used to describe the names of fabrics, colors, and styles of garments, which makes it difficult for the contemporary reader to understand extant or historical written material and interpret the attributes of the garment/artifact described accurately \cite{tortora_phyllis_survey_2015}. 
Dress artifacts are forms of primary data that provide ``evidence'' of what was worn and is considered ``more democratic'' than the written words \cite{prown_mind_1982}. 

In the context of historic clothing and textile collections, donors often provide inaccurate written descriptions of artifacts based on their limited knowledge of fashion terminology. Collection managers also adopt these inaccurate descriptions or may unintentionally misdescribe dress artifacts because the artifacts may have been remade or reworked over time. 
In such instances, the written word cannot be relied upon and a close study and examination of artifacts is necessary.

It is, therefore, a challenge to map a garment written description to a controlled vocabulary, specifically Costume Core. Costume Core is a metadata schema with controlled descriptive fashion terminology (vocabulary) for garments and accessories \cite{kirkland_costume_2018,kirkland_sharing_2016}. Cataloging using a controlled vocabulary is important as it minimizes any potential bias or incorrect interpretation from the curator as described above. A garment contains multiple attributes like Color, Medium, Work Type, Sleeve Type, etc. that describe it, each of which has a term in the Costume Core vocabulary. Having this task done via a human subject is obviously a tedious task. Natural Language Processing (NLP) techniques can be applied to map the historically written descriptions to the Costume Core vocabulary. 


In this paper, we build an NLP model to automatically map the free-form written garment descriptions to controlled vocabulary terms. We catalogued hundreds of garments and accessories along with their accompanying historic descriptions, and used them to train our NLP model. 

Having a relatively small number of descriptions was a challenge. The use of multiple terms of the same costume core category in the same description (e.g., multiple secondary colors in addition to the primary one) was also misleading to our model that we needed to go through iterations of model design and tuning to overcome. Despite those challenges, we were able to achieve above 90\% test accuracy.

While mapping historical artifacts to a controlled vocabulary has been traditionally done manually \cite{taylor_doing_1998}, automating the process is largely unexplored. The contribution of this paper is to introduce a level of automation into this process using NLP to save collection staff time and ultimately encourage the digitization of costume collections and increase their accessibility and use \cite{marcketti2019should}.


The rest of the paper is organized as follows. A brief literature review is provided in Section \ref{lit_review}. Section \ref{methodology} describes our materials and methodology with Sections \ref{experiments} and \ref{results} describing our experimental design and choices as well as the corresponding results. We provide a discussion of the results in Section \ref{discussion} followed by our conclusions in Section \ref{conclusion}.






\section{Literature Review}
\label{lit_review}

The study of dress (the process and product of dressing the body \cite{eicher_visible_2014}) is important to the humanities. Not only is dress a process in which most, if not all individuals, participate, but it is also an interdisciplinary field of study that draws from academic disciplines within the humanities such as anthropology, history, sociology, and psychology, which spark interests among a variety of scholars and students who contribute to humanistic scholarship. Historic costume artifacts are forms of primary data that provide “evidence” of what was worn in the past \cite{prown_mind_1982}. Thus, the artifacts are forms of material culture that indicate the social norms, values, mores, and ways of life that humans adopted during a specific time and place. More specifically, historic dress artifacts provide information about social structures regarding age, gender, social class, and race/ethnicity. For this reason, Costume Core provides controlled vocabularies relating to age/stage of life, gender, social class, and country of origin \cite{kirkland_costume_2018}. This assists in providing a record of the social and historical context behind each artifact that more clearly allows curators to examine and, in turn, communicate the cultural significance of dress and fashion to the public through exhibitions and online digital libraries.

Historically, the process of describing these artifacts using a controlled vocabulary, such as Costume Core, has been manually completed by close examination of each artifact \cite{taylor_doing_1998}. In this process, the accompanying description may or may not be used. It could be used to provide some clarifications on things such as material or fabrication. We view the accompanying descriptions as being a rich source of information that NLP techniques can use to aid in the describing pipeline. In essence, NLP techniques have the potential to assist in building accessible descriptions of these dress artifacts.


The problem addressed in this paper closely relates to text classification of which there are a number of approaches. The authors of \cite{kowsari_text_2019} perform a very detailed survey of techniques and approaches. Some researchers have investigated the use of different variations of recurrent neural networks \cite{lai_recurrent_2015,li_news_2018} where news articles are classified into different story types. However, with the arrival of a transformer encoder-decoder architecture \cite{vaswani_attention_2017} provided new opportunities for solving text classification tasks such as \cite{zheng_new_2019,li_automatic_2019}.

Focusing in on the specific problem presented in this paper, there has been work on training text classification on small datasets with multi-labels. Such work address feature selection methods for creating text classification solutions \cite{dzisevic_text_2019,kou_evaluation_2020}. The small size of our dataset leads us to the need of data augmentation. There are a number of techniques for text data augmentation \cite{shorten_text_2021}. However, due to the nature of our data and its conceptual meaning, we decided that back-translation \cite{sennrich_improving_2015} is the most applicable technique (more details in Section \ref{methodology}).

Since describing garments and accessories has been a manual process, to the authors' knowledge, there has not been any other work to date focused on automated multi-label classification of these items based on provided text descriptions.

\section{Materials and Methodology}
\label{methodology}

In this work, we followed the well-known NLP encoder-decoder model architecture where an encoder consumes the artifact description and produces a fixed length embedding vector that is used by a decoder to do the classification task. We describe below how we did this in more detail.

\subsection{Dataset}
The dataset we are working with stems from a curated costume and textile collection [citation withheld for blind review]. The collection consists of thousands of garments and accessories, however, only a subset has been labeled using Costume Core (380). This is because, as mentioned earlier, describing items (e.g., with Costume Core), is a time-consuming process performed manually. This describing process is currently being accomplished by the curator's team. 

As this is a curated collection from donors, each item is accompanied by a free-form text description provided by the donors. Each description was originally typed on a physical notecard. The notecards were digitized with optical character recognition (OCR) performed on the images to extract the text. This OCR'ed text was then inspected for quality control. The OCR technology used was Spark OCR \cite{noauthor_spark_nodate}. From several attributes in the controlled vocabulary, we focused on Color and Work Type as they apply to all garments and artifacts and are independent. We have developed a pipeline that can be later expanded for the other attributes. 


\subsection{The Learning Model}
There are two approaches towards building a model capable of processing complex input (text in our case): train a model from scratch or use transfer learning. In transfer learning, a model pre-trained on a large amount of similar data is connected to a small Feed Forward Neural Network (FNN). This structure harnesses the knowledge learned by the pre-trained model and fits this knowledge to the domain problem through training the connected FNN. Since we have a limited dataset, using transfer learning was our best option. To realize this, we used Google's Universal Sentence Encoder (USE) \cite{CerEtal2018_universal_sentence_encoder} to fill in the encoder part of our model representing the context of the input text. The USE reduces the knowledge contained in a sentence to a fixed length embedding vector. This is achieved through a transformer-based encoder, or a Deep Averaging Network based encoder. We used the transformer-based encoder which has a higher accuracy but is more complex and resource-heavy. The produced embeddings are fed to the FNN to proceed in the classification task .

\subsection{Data Augmentation}
While the use of transfer learning relatively reduces our need for data, we recognize that 380 garments descriptions are still not enough to produce a well-trained model. The lack of training data is a common problem in training machine learning models and can limit the model's ability to generalize well. To overcome the training data problem, data augmentation can be used to synthetically enrich our training data \cite{sinha2019data, rizk2018cellindeep}. We used back-translation to achieve this task, in which we translate a description to multiple different languages; namely French, German, and Spanish, then back to English. Despite the historic nature of the original text, the resulting different English versions of the description add a level of resiliency to our model and made it more generalizable. The size of the dataset after augmentation is 1520, which after sentence tokenization increases to 4169. 

\subsection{Evaluation}
\label{evaluation}
In machine learning, the training dataset should represent the original dataset distribution. The test set, however, plays the role of the unseen data, hence should not be impacted by the design choices to guarantee a fair evaluation of the trained model. While data augmentation enriches our dataset by producing sentences different from the original ones, the derived sentences can still exhibit some correlation to their original ones. Consequently, we needed to carefully split our data and evaluate the model predictions.

\begin{itemize}
    \item \textbf{Data Split}: Having correlated descriptions between the training set and test set can mislead the model evaluation scores by overestimating the model's ability to predict in a real scenario. To avert this effect, we split our dataset then augmented the training descriptions separately using the described augmentation techniques.
    \item \textbf{Prediction}: The augmentation techniques we used enriched our dataset with different versions of the original descriptions. Therefore, we evaluated our model predictability on each description in the test set by aggregating and averaging the prediction score across a set of descriptions produced via the same augmentation techniques used on the training set. This method was followed by some of the major seminal papers that were produced based on the ImageNet Large Scale Visual Recognition Challenge (ILSVRC) \cite{ILSVRC}, such as ResNet \cite{resent} and GoogLeNet \cite{GoogLeNet}.
\end{itemize}

In addition to using the average prediction score, and since our classification problem includes more than 10 labels, we used the top-$k$ error evaluation technique used in ILSVRC \cite{ILSVRC}. The top-$k$ error measures the possibility that the target label does not exist in the top $k$ predicted labels ($k$ labels with highest probability where $\sum{P_{class}}=1$). We used the top-$k$ error for $k$ up to 3. 

\section{Experiments}
\label{experiments}
\subsection{Intuition}
While a description contains multiple sentences describing the artifact in detail, information relevant to the primary color or Work Type is often present only in one sentence, and sometimes it includes multiple colors for different parts of the garment. Only one of them is considered the ``primary color'' and another is considered a ``secondary color'' and the rest are of no significance. For example, in the description ``Short brown, grey, beige mink fur cape. ", brown is the primary color, grey is the secondary color, and beige is of no significance. 

Irrelevant information potentially adds noise to the data and hinders the model from focusing on the right features. In some extreme cases, this irrelevant information can turn adversarial. An example of an adversarial case is when multiple color variants of the same artifact are mentioned in the description such as "robe: floorlength hot pink wool robe. crew neck lined with a brocade teal and gold ribbon going down the front". The robe is pink while the brocade is teal and ribbon is gold. 

At an early stage of this work, our model took the entire description as input. But this model was unable to generalize well given a multi-sentence description and a fixed length embedding, the model could not attend enough on the right context to capture the correct Costume Core Vocabulary.  
  
To fix this, we changed the structure of the descriptions and the grouping of the class labels. The intuition is to introduce context isolation and allow the model to better attend to the important details . 

\subsection{Changing Structure of Input Text}

To better attend to context, we tokenize the descriptions into multiple sentences. After sentence tokenization, each training data sample contains one sentence from a description rather than the whole description. This can help the model to focus on sentence-level context rather than a description-level context. Now that the data is on a sentence-level we re-annotated each sentence based on the information that the sentence contains. This is helpful in cases where there are multiple attribute values for one garment such as ``a \textit{pink} dress with \textit{white} detailing''. The model would classify every sentence into the corresponding primary color or work type of the original garment description if it is present in the sentence. Otherwise, a no-color/no-work-type label is attached to the sentence. Finally, data augmentation using back-translation is applied on the sentence-level. 

\subsection{Color Grouping}
Having many classes resulted in data sparsity class-wise. By reducing the number of classes, we were able to tackle this problem and ensure that there is enough data per class. The Costume Core Vocabulary defines 31 colors, which include metallic colors like rust, gold, silver for jewellery and accessories. These 31 classes were mapped to a broader grouping of colors as shown in Table \ref{tab:color-groups}. 

\begin{table}[]
\centering
\caption{Color Groups}
\label{tab:color-groups}
\begin{tabular}{|l|l|l|l|}
\hline
\textbf{color} & \textbf{color-group}      & \textbf{color} & \textbf{color-group}    \\ \hline
black          & black                     & coral          & \multirow{3}{*}{orange} \\ \cline{1-3}
blue           & \multirow{3}{*}{blue}     & orange         &                         \\ \cline{1-1} \cline{3-3}
navy blue      &                           & brass          &                         \\ \cline{1-1} \cline{3-4} 
teal           &                           & fuchsia        & \multirow{2}{*}{pink}   \\ \cline{1-3}
brown          & \multirow{2}{*}{brown}    & pink           &                         \\ \cline{1-1} \cline{3-4} 
tan            &                           & lavender       & \multirow{2}{*}{purple} \\ \cline{1-3}
gray           & \multirow{2}{*}{gray}     & purple         &                         \\ \cline{1-1} \cline{3-4} 
silver         &                           & burgundy       & \multirow{4}{*}{red}    \\ \cline{1-3}
green          & \multirow{2}{*}{green}    & maroon         &                         \\ \cline{1-1} \cline{3-3}
turquoise      &                           & red            &                         \\ \cline{1-3}
gold           & \multirow{3}{*}{metallic} & rust           &                         \\ \cline{1-1} \cline{3-4} 
metallic       &                           & beige          & \multirow{4}{*}{white}  \\ \cline{1-1} \cline{3-3}
gold           &                           & cream          &                         \\ \cline{1-3}
yellow         & \multirow{2}{*}{yellow}   & white          &                         \\ \cline{1-1} \cline{3-3}
amber          &                           & clear          &                         \\ \hline
\end{tabular}
\vspace*{-0.15in}
\end{table}

\subsection{Data Balancing}

\begin{figure}[htp]
    \centering
    \includegraphics[width=8cm]{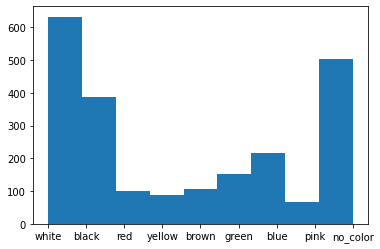}
    \vspace*{-0.15in}
    \caption{Class-wise Distribution of Training Data - Color}
    \vspace*{-0.15in}
    \label{fig:color-train-dist}
\end{figure}

\begin{figure}[htp]
    \centering
    \includegraphics[width=8cm]{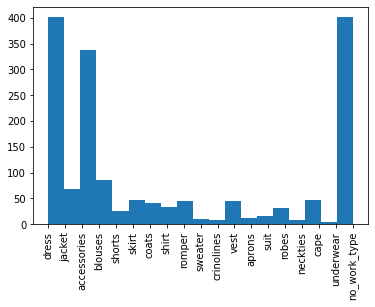}
    \vspace*{-0.15in}
    \caption{Class-wise Distribution of Training Data - Work Type}
    \vspace*{-0.15in}
    \label{fig:wt-train-dist}
\end{figure}

While our new labeling scheme improves the model attention, it introduces another challenge, data imbalance. After sentence tokenization, the data contains many more sentences labeled ``no-color" than those with a primary color. To counter this imbalance, undersampling is performed for the ``no-color" and ``no-work-type" class. The class is sampled by a fraction of 0.15 to match the largest classes like white and black as shown in Figures \ref{fig:color-train-dist} and \ref{fig:wt-train-dist}. By reducing to the largest class, we reduce the chance of model bias towards the majority class, ``no-color'' and ``no-work-type". The majority classes could not be further undersampled to obtain better data balance as that would significantly reduce the size of the dataset. It is worth mentioning that we apply data balancing only to the training data after shuffling and splitting the original dataset. As pointed out in Section \ref{evaluation}, each description in the test set is tokenized and augmented separately but without balancing.

\subsection{Architecture and Hyperparameters}
\begin{figure}[htp]
    \centering
    \vspace*{-0.2in}
    \includegraphics[width=8cm]{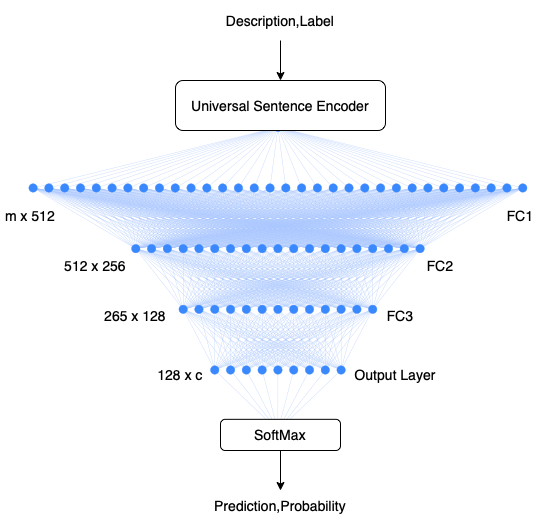}
    \vspace*{-0.15in}
    \caption{Model Architecture - A labeled description is provided to the USE to produce an embedding of size m. This embedding is then passed to the FC layers to obtain an output vector of size where c is the number of classes. The output vector is provided to a SoftMax layer which results in the prediction and probability of the prediction. }
    \label{fig:model-arch}
\end{figure}

Our model architecture consists of the USE and multiple FNN layers, as shown in Figure \ref{fig:model-arch}. The embedding is produced by the USE and then passed to a Fully Connected (FC) Layer of the FNN. The FNN consists of 3 Fully Connected Layers. The output of the final layer is passed through a SoftMax layer to produce the predictions and their corresponding probabilities.
The best configuration of hyperparameters that lead to the results in Section \ref{results} are summarized in Table \ref{tab:hyperparameters}. The values were tuned using grid search for a range of 1e-05 to 1e-02 for Learning rate and 4 to 128 for Batch Size. We experimented with Adam, SGD and RMSProp for optimizers. Gradient clipping was employed in experiments with SGD and RMSProp although Adam finally proved the best. Although we set the range to 20 epochs, we used early stopping to ensure that the model does not overfit. The raw dataset is split into 80/20 for the training and test sets. The training set and test set are augmented, and sentence tokenized separately. The processed training set is then further split into 80/20 as training and validation sets. 

\begin{table}
\centering
\caption{Hyperparameters of the model}
\label{tab:hyperparameters}
\begin{tabular}{|l|l|}
\hline
\textbf{Hyperparameter}             & \textbf{Value}              \\ \hline
Batch Size                 & 8                  \\ \hline
Learning Rate              & 0.001              \\ \hline
Optimizer                  & Adam               \\ \hline
$\beta_1$                  & 0.9                \\ \hline
$\beta_2$                  & 0.99               \\ \hline
$\varepsilon$               & 1e-07              \\ \hline
No. of epochs              & 20                 \\ \hline
Loss function               & Cross Entropy Loss \\ \hline
\end{tabular}
\vspace*{-0.2in}
\end{table}

\section{Results}
\label{results}

\subsection{Training and Validation Accuracy}
\begin{figure}[htp]
    \centering
    \includegraphics[width=6cm,height=3.65cm]{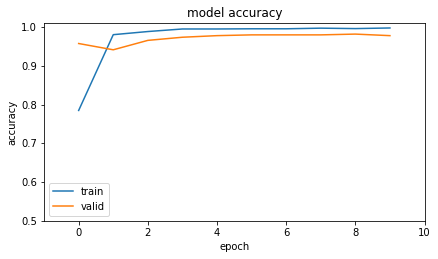}
    \vspace*{-0.15in}
    \caption{Training and Validation Accuracy for Color}
    \label{fig:color_train_val_acc_loss}
\end{figure}

\begin{figure}[htp]
    \centering
    \includegraphics[width=6cm,height=3.65cm]{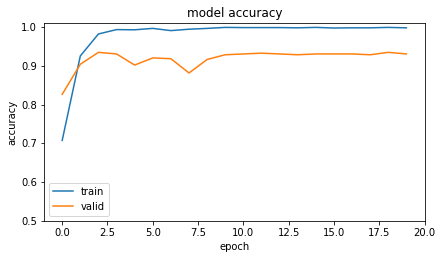}
    \vspace*{-0.15in}
    \caption{Training and Validation Accuracy for Work Type}
    \label{fig:wt_train_val_acc_loss}
\end{figure}

Figures \ref{fig:color_train_val_acc_loss} and \ref{fig:wt_train_val_acc_loss} show the model learning curves, particularly accuracy. For the color classification task, the model achieved training and validation accuracy of 99\% and 97.95\%, respectively. For the work type, the achieved training and validation accuracy are 99\% and 93.05\% respectively. The training and validation accuracy differ marginally. From the training accuracy we can infer that the model is learning the features of the data and the validation accuracy shows that it is learning it well. Based on the difference between the training and validation accuracy we can infer that the model does not overfit. Table \ref{tab:train-eg} shows an example of a description from the training dataset undergoing tokenization and augmentation before feeding to the model for training.

\begin{table}[]
\vspace*{-0.15in}
\centering
\caption{Example from Training Set}
\label{tab:train-eg}
\begin{tabular}{|l|l|}
\hline
Original Description &
  \begin{tabular}[c]{@{}l@{}}White and cream formal dress.\\ Fully covered in netting and lace.\\Cream taffeta, white netting with  \\ cream flocked and floral design.\end{tabular} \\ \hline
Example sentence after tokenization & white and cream formal dress      \\ \hline
Variants after augmentation &
  \begin{tabular}[c]{@{}l@{}}formal white dress and cream, \\ white and cream formal dress,\\ white dress and cream, \\ white and cream formal dress.\end{tabular} \\ \hline
Class label for Color               & white, no color, white            \\ \hline
Class label for Work Type           & dress,no work type,no work type \\ \hline
\end{tabular}
\vspace*{-0.15in}
\end{table}
 
\subsection{Test Results}
\begin{figure}[htp]
    \centering
    \vspace*{-0.1in}
    \includegraphics[width=8cm]{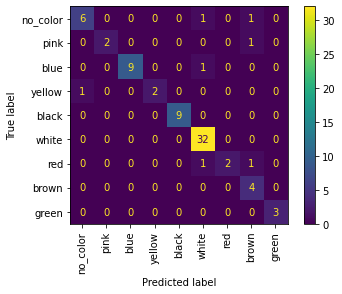}
    \vspace*{-0.15in}
    \caption{Confusion Matrix - Color}
    \label{fig:conf_mat}
\end{figure}

\begin{figure}[htp]
    \centering
    \vspace*{-0.15in}
    \includegraphics[width=8cm]{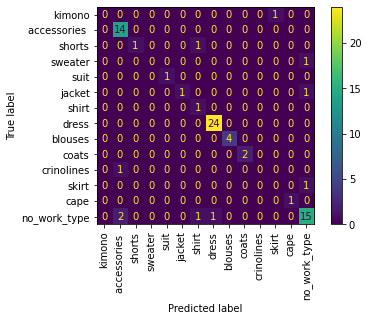}
    \vspace*{-0.15in}
    \caption{Confusion Matrix - Work Type}
    \vspace*{-0.15in}
    \label{fig:wt_conf_mat}
\end{figure}

The test set was separated from the training set and evaluated following the methodology described in Section \ref{methodology}. Tables \ref{tab:color_performance} and \ref{tab:wt_performance} show the overall evaluation metrics for the Color and Work Type attributes (precision, recall, and F1-score) as well as a breakdown for each color and Work Type class. Figures \ref{fig:conf_mat} and \ref{fig:wt_conf_mat} also show a detailed confusion matrix for the Color and Work Type attributes. The model achieves an F1-score of 0.91 for Color and 0.85 for Work Type. 

\begin{table}[]
\centering
\caption{Color Performance}
\label{tab:color_performance}
\begin{tabular}{|l|lll|l|}
\hline
\textbf{Color}    & \multicolumn{1}{l|}{\textbf{Precision}} & \multicolumn{1}{l|}{\textbf{Recall}} & \textbf{F1-Score} & \textbf{Support} \\ \hline
black    & \multicolumn{1}{l|}{1}         & \multicolumn{1}{l|}{1}      & 1        & 9       \\ \hline
blue     & \multicolumn{1}{l|}{1}         & \multicolumn{1}{l|}{0.9}    & 0.95     & 10      \\ \hline
brown    & \multicolumn{1}{l|}{0.57}      & \multicolumn{1}{l|}{1}      & 0.73     & 4       \\ \hline
green    & \multicolumn{1}{l|}{1}         & \multicolumn{1}{l|}{1}      & 1        & 3       \\ \hline
no color & \multicolumn{1}{l|}{0.86}      & \multicolumn{1}{l|}{0.75}   & 0.8      & 8       \\ \hline
pink     & \multicolumn{1}{l|}{1}         & \multicolumn{1}{l|}{0.67}   & 0.8      & 3       \\ \hline
red      & \multicolumn{1}{l|}{1}         & \multicolumn{1}{l|}{0.5}    & 0.67     & 4       \\ \hline
white    & \multicolumn{1}{l|}{0.91}      & \multicolumn{1}{l|}{1}      & 0.96     & 32      \\ \hline
yellow   & \multicolumn{1}{l|}{1}         & \multicolumn{1}{l|}{0.67}   & 0.8      & 3       \\ \hline
\multicolumn{3}{|l|}{accuracy}                                   & 0.91 & 76     \\ \hline
\end{tabular}
\vspace*{-0.15in}
\end{table}

\begin{figure*}
    \centering
    \includegraphics[width=18cm,height=3.75cm]{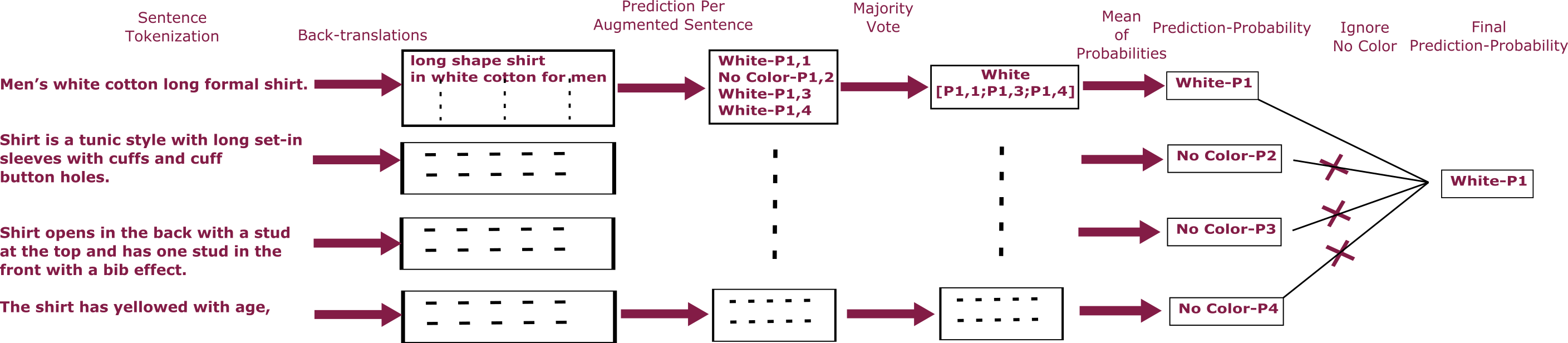}
    \caption{ Tokenization Pipeline: A tokenized sentence is aggregated with its augmented variants to produce a set of predictions and probabilities. The final prediction is then derived from majority vote and final probability from the mean of all probabilities.}
    \label{fig:test_results_pipeline}
    \vspace*{-0.15in}
\end{figure*}

\begin{table}[]
\vspace*{-0.25in}
\centering
\caption{Work Type Performance}
\label{tab:wt_performance}
\begin{tabular}{|lll|l|l|}
\hline
\multicolumn{1}{|l|}{\textbf{Work Type}}      & \multicolumn{1}{l|}{\textbf{Precision}} & \textbf{Recall} & \textbf{F1-score} & \textbf{Support} \\ \hline
\multicolumn{1}{|l|}{accessories}    & \multicolumn{1}{l|}{0.82}      & 1.00   & 0.90     & 14      \\ \hline
\multicolumn{1}{|l|}{blouses} & \multicolumn{1}{l|}{1.00} & 1.00 & 1.00 & 4  \\ \hline
\multicolumn{1}{|l|}{cape}    & \multicolumn{1}{l|}{1.00} & 1.00 & 1.00 & 1  \\ \hline
\multicolumn{1}{|l|}{coats}   & \multicolumn{1}{l|}{1.00} & 1.00 & 1.00 & 2  \\ \hline
\multicolumn{1}{|l|}{crinolines}     & \multicolumn{1}{l|}{0.00}      & 0.00   & 0.00     & 1       \\ \hline
\multicolumn{1}{|l|}{dress}   & \multicolumn{1}{l|}{0.96} & 1.00 & 0.98 & 24 \\ \hline
\multicolumn{1}{|l|}{jacket}  & \multicolumn{1}{l|}{1.00} & 0.50 & 0.67 & 2  \\ \hline
\multicolumn{1}{|l|}{kimono}  & \multicolumn{1}{l|}{0.00} & 0.00 & 0.00 & 1  \\ \hline
\multicolumn{1}{|l|}{no\_work\_type} & \multicolumn{1}{l|}{0.83}      & 0.79   & 0.81     & 19      \\ \hline
\multicolumn{1}{|l|}{shirt}   & \multicolumn{1}{l|}{0.33} & 1.00 & 0.50 & 1  \\ \hline
\multicolumn{1}{|l|}{shorts}  & \multicolumn{1}{l|}{1.00} & 0.50 & 0.67 & 2  \\ \hline
\multicolumn{1}{|l|}{suit}    & \multicolumn{1}{l|}{1.00} & 1.00 & 1.00 & 1  \\ \hline
\multicolumn{1}{|l|}{sweater} & \multicolumn{1}{l|}{0.00} & 0.00 & 0.00 & 2  \\ \hline
\multicolumn{3}{|l|}{accuracy}                                   & 0.85 & 75 \\ \hline
\end{tabular}

\end{table}
Figure \ref{fig:test_results_pipeline} shows a description from a test record for which predictions have been obtained by the model. It tracks a description before aggregation on a sentence-level to the final prediction on a description level as described in Section \ref{methodology}. 

\begin{table}[]
\centering
\caption{Results}
\label{tab:final-results}
\begin{tabular}{|l|l|l|}
\hline
\textbf{Experiment}            & \textbf{Evaluation}                                                         & \textbf{Test Accuracy(\%)} \\ \hline
\begin{tabular}[c]{@{}l@{}}Initial approach of testing \\ on original descriptions\end{tabular} &
  \begin{tabular}[c]{@{}l@{}}Standard \\ accuracy \\ evaluation\end{tabular} &
  44 \\ \hline
\begin{tabular}[c]{@{}l@{}}Augmentation \& \\ Color Grouping\end{tabular} &
  \begin{tabular}[c]{@{}l@{}}Standard \\ accuracy \\ evaluation\end{tabular} &
  52 \\ \hline
Testset augmentation  & \begin{tabular}[c]{@{}l@{}}Top-k \\ accuracy\end{tabular}          & 78 for top-3      \\ \hline
Sentence Tokenization & \begin{tabular}[c]{@{}l@{}}Our evaluation \\ strategy\end{tabular} & 91                \\ \hline
\end{tabular}
\vspace*{-0.15in}
\end{table}

\section{Discussion}
\label{discussion}


When working with detailed text descriptions where each sentence describes a different part of the garment, it is important to capture the relevant features to achieve good performance, i.e., the model needs to attend to the right context. In the English language, context is critical as one word can have many meanings. In our task of Color classification, a model cannot distinguish between the primary and secondary color unless the importance is induced through context.  
  
We started by training the model on the descriptions after preprocessing. We achieved a high training and validation accuracy over 95\%, but the test accuracy was 44\%. The model was unable to generalize well. In a complex training record, we see that multiple details are mentioned in the description making it difficult for the model to learn the right feature. While analyzing the results, we observed the accuracy increase from 44\% to 63\% if we accepted a secondary color in the prediction as the correct prediction. The model had trouble capturing the key features. We suspected that the model was overfitting since the model complexity was high while the dataset was small and sparse.  
  
Our next effort was to provide more examples per class to the model through augmentation, hoping that would reinforce the important features and make the model more robust. Since we had already achieved a good accuracy on training and validation, we focused on the generalization capabilities of the model which can be measured through the test accuracy. Although that helped in improving accuracy by 8\%, reaching 52\%, it was not good enough. When we evaluated the model with top-k accuracy as described in Section \ref{methodology}, we saw an increase in test accuracy to 78\% for the top-3 accuracy. We observed that as we increased $k$, the accuracy increased and then saturated at $k = 3$. Observing that the primary color was in the top-3 predictions of the model hinted that the model requires more attention for the primary class, so that it could appear as the top prediction instead of one amongst the top-3.  
  
To provide the model with the right context to attend to, we simplified the description and provided only sentences (fragments of a description) to the model to train on. By sentence tokenization, we were able to break down the complexity associated with multiple sentences where each sentence referred to at least one other attribute of the garment. This helped the model concentrate more on the class present in the given sentence. But colors may not appear in every sentence of the model. Since we wanted our model to be robust to descriptions where there might be no color cue present, we labeled the sentences as color $c$ if it were present as the primary color, otherwise it was labeled ``no-color''. Now the model not only could identify colors in each sentence, but it could also identify sentences that do not have a primary color in them. By breaking down the sentences, we were able to also capture different colors in a garment if they appeared in different sentences. As an example, in the description "Girl's pink organza dress. Dress has a peter pan collar trimmed in lace and floral embroidery in white", by tokenizing it into multiple sentences, the first sentence is labeled ``pink'' while the rest are ``no-color''. Now, we can label the second sentence as white in case we wished to capture secondary color in the future. The results are summarized in Table \ref{tab:final-results}. 
  
The final goal of the work is to be able to automatically map a new set of descriptions to their Costume Core vocabulary. If the model were to give a sentence-level result when provided with a description, there would be uncertainty while using the resulting predictions for further tasks. Hence, we used aggregation for the model predictions. Through aggregation, we get one final prediction for a description that can be traced back to the sentence-level results if required. If the requirement were to change to a sentence-level prediction or to identify if color is present, the results can be used as such or aggregated however deemed fit. This flexibility of using a sentence or description-level prediction enables the model to be used with different tasks. The results can be used in conjunction with the confusion matrix to interpret and understand why a particular class is performing well or poorly. For color, we can see that white is performing well with a precision of 0.91 and a recall of 1. A recall score of 1 means that all the 32 white descriptions were classified correctly as white. But a lower precision score indicates that there may be false positives for the class. Looking at the confusion matrix, we can see that one data sample, which did not have any primary color, was misclassified as white.  


\section{Conclusion and Future Work}
\label{conclusion}
In this paper we use machine learning to automate the Costume Core classification problem for description of historic garments, and we particularly target the garment Primary Color and Work Type. In this problem, we seek to extract these two attributes from the garment accompanying free-form text descriptions, where a human expert would otherwise be needed. We present an NLP solution based on the USE. In the presented approach, we described our methodology including the preprocessing and data augmentation needed to get the garment descriptions ready for consumption by the model. We also described how the evaluation method is framed to follow the rationale of the steps taken in training the model. While our early trials had less than 50\% test accuracy, we managed to adjust and tune the model to reach more than 90\% accuracy. Given a small dataset (380 garment descriptions), we consider the accuracy achieved acceptable and promising. As there are more descriptions currently being OCR'ed and prepared for use, we anticipate having this accuracy increase. This solution can be further extended to other Costume Core attributes such as Medium, Technique and Dress Type. As a part of the collection, some artifacts are also photographed. The images can be used in addition to the descriptions and a multimodal machine learning model can be built on this dataset in the future.  

\section{Acknowledgements}
\label{acknowledgement}
This work is partially supported by the Virginia Tech Libraries Collaborative Research Grant (January 2022).

\bibliographystyle{IEEEtran}
\bibliography{lit}

\begin{thebibliography}{10}
\providecommand{\url}[1]{#1}
\csname url@samestyle\endcsname
\providecommand{\newblock}{\relax}
\providecommand{\bibinfo}[2]{#2}
\providecommand{\BIBentrySTDinterwordspacing}{\spaceskip=0pt\relax}
\providecommand{\BIBentryALTinterwordstretchfactor}{4}
\providecommand{\BIBentryALTinterwordspacing}{\spaceskip=\fontdimen2\font plus
\BIBentryALTinterwordstretchfactor\fontdimen3\font minus
  \fontdimen4\font\relax}
\providecommand{\BIBforeignlanguage}[2]{{%
\expandafter\ifx\csname l@#1\endcsname\relax
\typeout{** WARNING: IEEEtran.bst: No hyphenation pattern has been}%
\typeout{** loaded for the language `#1'. Using the pattern for}%
\typeout{** the default language instead.}%
\else
\language=\csname l@#1\endcsname
\fi
#2}}
\providecommand{\BIBdecl}{\relax}
\BIBdecl

\bibitem{tortora_phyllis_survey_2015}
G.~Tortora~Phyllis and B.~Marcketti~Sara, \emph{Survey of {Historic}
  {Costume}}, 6th~ed.\hskip 1em plus 0.5em minus 0.4em\relax Fairchild Books,
  2015.

\bibitem{prown_mind_1982}
\BIBentryALTinterwordspacing
J.~D. Prown, ``\BIBforeignlanguage{en}{Mind in {Matter}: {An} {Introduction} to
  {Material} {Culture} {Theory} and {Method}},''
  \emph{\BIBforeignlanguage{en}{Winterthur Portfolio}}, vol.~17, no.~1, pp.
  1--19, Apr. 1982. [Online]. Available:
  \url{https://www.journals.uchicago.edu/doi/10.1086/496065}
\BIBentrySTDinterwordspacing

\bibitem{kirkland_costume_2018}
\BIBentryALTinterwordspacing
A.~Kirkland, ``\BIBforeignlanguage{en}{Costume {Core}: {Metadata} for
  {Historic} {Clothing}},'' \emph{\BIBforeignlanguage{en}{Visual Resources
  Association Bulletin}}, vol.~45, no.~2, 2018, number: 2. [Online]. Available:
  \url{https://online.vraweb.org/index.php/vrab/article/view/36}
\BIBentrySTDinterwordspacing

\bibitem{kirkland_sharing_2016}
\BIBentryALTinterwordspacing
A.~Kirkland, K.~Martin, M.~Schoeny, K.~Smith, and G.~Strege,
  ``\BIBforeignlanguage{en}{Sharing {Historic} {Costume} {Collections}
  {Online}: {Why} and {How}},'' \emph{\BIBforeignlanguage{en}{Dress}}, Jan.
  2016, publisher: Routledge. [Online]. Available:
  \url{https://www.tandfonline.com/doi/abs/10.1080/03612112.2015.1130394}
\BIBentrySTDinterwordspacing

\bibitem{taylor_doing_1998}
\BIBentryALTinterwordspacing
L.~Taylor, ``Doing the {Laundry}? {A} {Reassessment} of {Object}-based {Dress}
  {History},'' \emph{Fashion Theory}, vol.~2, no.~4, pp. 337--358, Nov. 1998,
  publisher: Routledge \_eprint: https://doi.org/10.2752/136270498779476118.
  [Online]. Available: \url{https://doi.org/10.2752/136270498779476118}
\BIBentrySTDinterwordspacing

\bibitem{marcketti2019should}
S.~Marcketti and J.~F. Gordon, ``I should probably know more: Reasons for and
  roadblocks to the use of historic university collections in teaching,''
  \emph{Journal of Conservation and Museum Studies}, vol.~17, no.~1, 2019.

\bibitem{eicher_visible_2014}
J.~B. Eicher and S.~L. Evenson, \emph{\BIBforeignlanguage{en}{The {Visible}
  {Self}: {Global} {Perspectives} on {Dress}, {Culture} and {Society}}}.\hskip
  1em plus 0.5em minus 0.4em\relax Bloomsbury Publishing USA, Aug. 2014,
  google-Books-ID: X3qXBgAAQBAJ.

\bibitem{kowsari_text_2019}
\BIBentryALTinterwordspacing
K.~Kowsari, K.~Jafari~Meimandi, M.~Heidarysafa, S.~Mendu, L.~Barnes, and
  D.~Brown, ``\BIBforeignlanguage{en}{Text {Classification} {Algorithms}: {A}
  {Survey}},'' \emph{\BIBforeignlanguage{en}{Information}}, vol.~10, no.~4, p.
  150, Apr. 2019, number: 4 Publisher: Multidisciplinary Digital Publishing
  Institute. [Online]. Available: \url{https://www.mdpi.com/2078-2489/10/4/150}
\BIBentrySTDinterwordspacing

\bibitem{lai_recurrent_2015}
\BIBentryALTinterwordspacing
S.~Lai, L.~Xu, K.~Liu, and J.~Zhao, ``\BIBforeignlanguage{en}{Recurrent
  {Convolutional} {Neural} {Networks} for {Text} {Classification}},'' in
  \emph{\BIBforeignlanguage{en}{Twenty-{Ninth} {AAAI} {Conference} on
  {Artificial} {Intelligence}}}, Feb. 2015. [Online]. Available:
  \url{https://www.aaai.org/ocs/index.php/AAAI/AAAI15/paper/view/9745}
\BIBentrySTDinterwordspacing

\bibitem{li_news_2018}
\BIBentryALTinterwordspacing
C.~Li, G.~Zhan, and Z.~Li, ``News {Text} {Classification} {Based} on {Improved}
  {Bi}-{LSTM}-{CNN},'' in \emph{2018 9th {International} {Conference} on
  {Information} {Technology} in {Medicine} and {Education} ({ITME})}.\hskip 1em
  plus 0.5em minus 0.4em\relax Hangzhou: IEEE, Oct. 2018, pp. 890--893.
  [Online]. Available: \url{https://ieeexplore.ieee.org/document/8589431/}
\BIBentrySTDinterwordspacing

\bibitem{vaswani_attention_2017}
\BIBentryALTinterwordspacing
A.~Vaswani, N.~Shazeer, N.~Parmar, J.~Uszkoreit, L.~Jones, A.~N. Gomez,
  Å.~Kaiser, and I.~Polosukhin, ``\BIBforeignlanguage{en}{Attention is {All}
  you {Need}},'' \emph{\BIBforeignlanguage{en}{Advances in Neural Information
  Processing Systems}}, vol.~30, 2017. [Online]. Available:
  \url{https://proceedings.neurips.cc/paper/2017/hash/3f5ee243547dee91fbd053c1c4a845aa-Abstract.html}
\BIBentrySTDinterwordspacing

\bibitem{zheng_new_2019}
S.~Zheng and M.~Yang, ``\BIBforeignlanguage{en}{A {New} {Method} of {Improving}
  {BERT} for {Text} {Classification}},'' in
  \emph{\BIBforeignlanguage{en}{Intelligence {Science} and {Big} {Data}
  {Engineering}. {Big} {Data} and {Machine} {Learning}}}, ser. Lecture {Notes}
  in {Computer} {Science}, Z.~Cui, J.~Pan, S.~Zhang, L.~Xiao, and J.~Yang,
  Eds.\hskip 1em plus 0.5em minus 0.4em\relax Cham: Springer International
  Publishing, 2019, pp. 442--452.

\bibitem{li_automatic_2019}
\BIBentryALTinterwordspacing
W.~Li, S.~Gao, H.~Zhou, Z.~Huang, K.~Zhang, and W.~Li, ``The {Automatic} {Text}
  {Classification} {Method} {Based} on {BERT} and {Feature} {Union},'' in
  \emph{2019 {IEEE} 25th {International} {Conference} on {Parallel} and
  {Distributed} {Systems} ({ICPADS})}.\hskip 1em plus 0.5em minus 0.4em\relax
  Tianjin, China: IEEE, Dec. 2019, pp. 774--777. [Online]. Available:
  \url{https://ieeexplore.ieee.org/document/8975793/}
\BIBentrySTDinterwordspacing

\bibitem{dzisevic_text_2019}
\BIBentryALTinterwordspacing
R.~Dzisevic and D.~Sesok, ``Text {Classification} using {Different} {Feature}
  {Extraction} {Approaches},'' in \emph{2019 {Open} {Conference} of
  {Electrical}, {Electronic} and {Information} {Sciences} ({eStream})}.\hskip
  1em plus 0.5em minus 0.4em\relax Vilnius, Lithuania: IEEE, Apr., pp. 1--4.
  [Online]. Available: \url{https://ieeexplore.ieee.org/document/8732167/}
\BIBentrySTDinterwordspacing

\bibitem{kou_evaluation_2020}
\BIBentryALTinterwordspacing
G.~Kou, P.~Yang, Y.~Peng, F.~Xiao, Y.~Chen, and F.~E. Alsaadi,
  ``\BIBforeignlanguage{en}{Evaluation of feature selection methods for text
  classification with small datasets using multiple criteria decision-making
  methods},'' \emph{\BIBforeignlanguage{en}{Applied Soft Computing}}, vol.~86,
  p. 105836, Jan. 2020. [Online]. Available:
  \url{https://linkinghub.elsevier.com/retrieve/pii/S1568494619306179}
\BIBentrySTDinterwordspacing

\bibitem{shorten_text_2021}
\BIBentryALTinterwordspacing
C.~Shorten, T.~M. Khoshgoftaar, and B.~Furht, ``\BIBforeignlanguage{en}{Text
  {Data} {Augmentation} for {Deep} {Learning}},''
  \emph{\BIBforeignlanguage{en}{Journal of Big Data}}, vol.~8, no.~1, p. 101,
  Dec. 2021. [Online]. Available:
  \url{https://journalofbigdata.springeropen.com/articles/10.1186/s40537-021-00492-0}
\BIBentrySTDinterwordspacing

\bibitem{sennrich_improving_2015}
\BIBentryALTinterwordspacing
R.~Sennrich, B.~Haddow, and A.~Birch, ``Improving {Neural} {Machine}
  {Translation} {Models} with {Monolingual} {Data},'' 2015, publisher: arXiv
  Version Number: 4. [Online]. Available:
  \url{https://arxiv.org/abs/1511.06709}
\BIBentrySTDinterwordspacing

\bibitem{noauthor_spark_nodate}
\BIBentryALTinterwordspacing
``\BIBforeignlanguage{en-US}{Spark {OCR}}.'' [Online]. Available:
  \url{https://www.johnsnowlabs.com/spark-ocr/}
\BIBentrySTDinterwordspacing

\bibitem{CerEtal2018_universal_sentence_encoder}
\BIBentryALTinterwordspacing
D.~Cer, Y.~Yang, S.~Kong, N.~Hua, N.~Limtiaco, R.~S. John, N.~Constant,
  M.~Guajardo{-}Cespedes, S.~Yuan, C.~Tar, Y.~Sung, B.~Strope, and R.~Kurzweil,
  ``Universal sentence encoder,'' \emph{CoRR}, vol. abs/1803.11175, 2018.
  [Online]. Available: \url{http://arxiv.org/abs/1803.11175}
\BIBentrySTDinterwordspacing

\bibitem{sinha2019data}
R.~S. Sinha, S.-M. Lee, M.~Rim, and S.-H. Hwang, ``Data augmentation schemes
  for deep learning in an indoor positioning application,'' \emph{Electronics},
  vol.~8, no.~5, p. 554, 2019.

\bibitem{rizk2018cellindeep}
H.~Rizk, M.~Torki, and M.~Youssef, ``{CellinDeep: Robust and accurate
  cellular-based indoor localization via deep learning},'' \emph{IEEE Sensors
  Journal}, vol.~19, no.~6, pp. 2305--2312, 2018.

\bibitem{ILSVRC}
\BIBentryALTinterwordspacing
Imagenet large scale visual recognition challenge. [Online]. Available:
  \url{https://www.image-net.org/challenges/LSVRC/}
\BIBentrySTDinterwordspacing

\bibitem{resent}
K.~He, X.~Zhang, S.~Ren, and J.~Sun, ``Deep residual learning for image
  recognition,'' in \emph{Proceedings of the IEEE Conference on Computer Vision
  and Pattern Recognition (CVPR)}, June 2016.

\bibitem{GoogLeNet}
C.~Szegedy, W.~Liu, Y.~Jia, P.~Sermanet, S.~Reed, D.~Anguelov, D.~Erhan,
  V.~Vanhoucke, and A.~Rabinovich, ``Going deeper with convolutions,'' in
  \emph{Proceedings of the IEEE conference on computer vision and pattern
  recognition}, 2015, pp. 1--9.

\end{thebibliography}

\end{document}